%% file: deepfinance.tex
\documentclass[11pt]{article}

\usepackage[margin=1in]{geometry}
\usepackage{graphicx}
\usepackage{booktabs}
\usepackage{xcolor}
\usepackage{caption}
\usepackage{float}
\usepackage[section]{placeins}
\usepackage{url}
\usepackage[colorlinks=true,linkcolor=blue!60!black,citecolor=blue!60!black,urlcolor=blue!60!black]{hyperref}

\graphicspath{{figures/}}
\newcommand{\reasoning}[1]{\textcolor{gray}{(#1)}}

\title{DeepFinance: Can AI Diligence a \$100MM Deal?}
\author{%
  \textbf{Drew Hahn}\textsuperscript{*} \quad
  \textbf{Tyler Shibata}\textsuperscript{*} \quad
  \textbf{Jackson Grove}\textsuperscript{*} \quad
  \textbf{Rahul Sonwalkar}\textsuperscript{*}\\[4pt]
  Julius Research, San Francisco}
\date{November 5, 2026}

\begin{document}

\maketitle
{\renewcommand{\thefootnote}{*}\footnotetext{\textit{All contributors contributed equally.} Correspondence to: \href{mailto:research@julius.ai}{research@julius.ai}.}}

\begin{abstract}
Existing evaluations of AI models on high-finance work focus on retrieving answers from financial documents and producing numerically correct outputs. Real private equity (PE) and investment banking (IB) work also depends on choosing between conflicting sources, making sound assumptions, and keeping every step auditable. We introduce DeepFinance, an ultra long-horizon benchmark of 21 end-to-end PE and IB tasks that take a human professional an average of 27.1 hours. Each task pairs a request with a data room averaging 19 files and is graded against an average of 84 expert-written rubric criteria covering output, workflow, and judgment. We evaluate 22 model configurations, each in its native agent harness. The best configuration, GPT-6 Astra with max reasoning, satisfies 57.6\% of rubric criteria, and the top four configurations are not statistically distinguishable. Frontier models often calculate correctly from the wrong inputs and find decision-relevant evidence without acting on it.
\end{abstract}

\section{Introduction}
\label{sec:introduction}

DeepFinance is an ultra long-horizon benchmark for measuring models on real-world private equity (PE) and investment banking (IB) workflows that would take a human professional between 20 and 30 hours to complete. DeepFinance was composed and reviewed by PE and IB professionals from firms such as J.P.~Morgan, UBS, and Evercore, and institutions such as Harvard Business School, Wharton, and Columbia Business School. The benchmark has the following characteristics:

\begin{itemize}
  \item Task count: 21
  \item Average human time: 27.1 hours per task
  \item Average rubric criteria: 84 criteria per task
  \item Average environment size: 19 files per data room
  \item Average prompt length: 240 words, approximately 29\% shorter than GDPval~\cite{gdpval} prompts
\end{itemize}

We evaluate 22 model configurations from 12 labs. The highest mean rubric score is 57.6\%, and the four leading configurations are not statistically distinguishable (Section~\ref{sec:results}). The leaderboard and an interactive sample task are available at \url{https://www.juliusresearch.ai/blog/deep-finance}.

\section{Motivation}
\label{sec:motivation}

There is a fundamental gap in how AI models are evaluated on real-world high-finance work. We believe this gap stems from a deeply flawed understanding of how analysts in high finance perform real-world tasks.

Existing evaluations measure models in two ways: their ability to retrieve answers from a set of financial documents, and their ability to produce numerically correct final outputs. Real-world PE and IB workflows are meaningfully more complicated.

In software benchmarks such as SWE-bench~\cite{swebench} and DeepSWE~\cite{deepswe}, models are graded on whether their output passes tests. If the tests pass and the fix works, the path (or trajectory) the agent took matters less. Finance has no similar test suite. A high-finance valuation is like a pull request that has to pass a line-by-line code review, where every number cites its source and every assumption needs to hold up in front of the managing director, the investment committee (IC), and a client.

In PE and IB, two analysts could produce identical numbers for the same task while one output rests on completely incorrect assumptions. Change a single number upstream, and the entire workflow no longer makes sense. The trajectory of an ultra long-horizon high-finance task is full of assumptions and judgment calls. Without being able to audit the analyst's steps and judgment, a professional in high finance cannot blindly trust the work presented to them.

\section{Related Work}
\label{sec:related-work}

Existing high-finance benchmarks have two main limitations in how they evaluate models: (1) they overemphasize question answering over financial documents, and (2) their grading is limited to the numerical correctness of outputs.

Existing benchmarks also often fail to consider the working environment that high-finance professionals face daily. A set of financial documents and revisions in varied formats, long email chains, sparse meeting notes, and other data-room objects almost invariably contains conflicting information that must be reconciled. A model's ability to judge and choose between overlapping sources is rarely measured.

\begin{itemize}
  \item \textbf{FinanceBench}~\cite{financebench}: questions are simple and too easy, and are not realistic or representative of the deal work analysts are hired for.
  \item \textbf{BizBench}~\cite{bizbench}: checks one final number against an answer key; DeepFinance grades end-to-end workflows against up to 140 criteria.
  \item \textbf{FinanceQA}~\cite{financeqa}: supplies the relevant 10-K sections with each question, leaving no room to evaluate the model's trajectory.
  \item \textbf{BigFinanceBench}~\cite{bigfinancebench}: uses solid trajectory reasoning but limits the scope of its questions and stops well short of end-to-end workflows.
\end{itemize}

Benchmarks such as OfficeQA~\cite{officeqa} track a company taxonomy well, but rarely go beyond grading numerical correctness. For ultra long-horizon high-finance tasks, the trajectory is crucial for understanding and auditing the method and assumptions behind the work. ELO-based preference ranking, as used by GDPval~\cite{gdpval}, compares only final outputs, misses the nuance of complex long-horizon workflows, and does not grade the atomic trajectory.

Some benchmarks, such as Halluminate's Westworld Finance Diligence Bench~\cite{westworld}, have attempted to measure semi-long-horizon work. They study the agent's trajectory but leave it out of the score: every point on their leaderboard comes from checks on the final output. This may be an effective way to check final answers, but it is not how finance work is judged in the real world. Correct numbers without a source or auditability will be scrutinized by an investment committee. DeepFinance grades whether every number is sourced and every judgment call is auditable and traceable.

DeepFinance was built on the premise that the path of a task carries its assumptions, and that grading must therefore account for the trajectory. Auditability is paramount to internal and external stakeholders such as lenders, clients, and brokers. Understanding how a model (or an analyst) arrived at an assumption is key to understanding what could go right, wrong, or sideways in a multi-million-dollar deal. Table~\ref{tab:criteria} compares rubric granularity across benchmarks.

\begin{table}[htbp]
  \centering
  \caption{Average number of rubric criteria per task.}
  \label{tab:criteria}
  \begin{tabular}{lr}
    \toprule
    Benchmark & Criteria per task \\
    \midrule
    BigFinanceBench & 16.9 \\
    PRBench Finance & 17.1 \\
    BlueFin & 35.1 \\
    \textbf{DeepFinance} & \textbf{84} \\
    \bottomrule
  \end{tabular}
\end{table}

\section{Methodology}
\label{sec:methodology}

\subsection{Task Construction}
\label{sec:task-construction}

DeepFinance covers four primary pillars of high-finance work:

\begin{itemize}
  \item \textbf{Valuation:} estimate what a business is worth and justify the assumptions.
  \item \textbf{Due diligence:} investigate the records, identify important risks, and explain their effect on the deal.
  \item \textbf{Underwriting:} assess the deal's risks and economics, then determine acceptable financing or investment terms.
  \item \textbf{Financial modeling:} build a model whose assumptions, calculations, and outputs remain consistent.
\end{itemize}

Each ultra long-horizon task pairs a request with a data room: the collection of records needed to complete the assignment. Tasks use real data that has been scrubbed and anonymized. The request states the objective, constraints, and expected deliverables; from there, the agent must work out how to complete the analysis. Problem statements average 240 words, which we find realistic for an end-to-end workflow. Other evaluations use much shorter prompts, but cover only simple arithmetic or quantitative tasks.

Each data room contains on average 19 files and comprises the same material analysts work with day to day. File types include spreadsheets (.xlsx), Word documents (.docx), emails (.eml), PDFs, and raw CSV extracts. Contents include emails, chats, and deal-team notes; models and projections; CIMs, SPAs, term sheets, commitment letters, and transaction data; diligence reports; IC minutes; and market data and comparables.

Each task is representative of end-to-end work at top PE and IB firms such as Evercore, J.P.~Morgan, and UBS. The tasks simulate analyst work on real deals, not simple question answering over a set of documents or the search for one-off answers. There is ambiguity, and there are conflicting sources.

\subsection{Sample Task}
\label{sec:sample-task}

The prompt for an investment due diligence task is reproduced below.

\begin{quote}
The IC is meeting this afternoon on Corliss.

Declan came back last night. He still wants \$50.50 at completion and another \$3.00 paid later if both Quarrie parts go into prod. What would we have to pay at completion and what would it cost us? What price would you put to IC? Show the split at that price and how you get to your numbers.

If IC agrees, signing is probably happening end of Jan. Thx, G
\end{quote}

The task comes with a data room of 18 files: a close ledger, supporting registers, a management pack, executed instruments, a pricing framework, a financial reporting policy, debt schedules for two peer companies, the share rights and option plan, a securities register, the sale mandate, an IC price check template, an earlier working, four emails, and an investment team chat export. To answer, the agent must reconcile FY2025 revenue and EBITDA, decide which customer programs count toward sustainable EBITDA under the pricing framework, value the company against its peers, and run the shareholder waterfall for both Declan's structure and an all-cash offer. The reference solution recommends offering 50.803 (USD millions) in cash with no Quarrie payment, with completion nominated for April 1, 2026. The task's rubric has 136 criteria. Appendix~\ref{app:sample-task} lists the data room files, the reference solution's key figures, and a sample of rubric criteria.

\subsection{Scoring}
\label{sec:scoring}

Graders evaluate the finished deliverable together with evidence of the trajectory taken. Sources, tool results, calculations, and revisions can show whether the agent completed the required work. Agents are graded against rubrics whose criteria are labeled with one of three categories:

\begin{itemize}
  \item \textbf{Output:} Are the calculations, conclusions, and requested deliverables correct?
  \item \textbf{Workflow:} Did the agent find the relevant evidence and use it in the analysis?
  \item \textbf{Judgment:} Does the evidence support the agent's assumptions and decisions?
\end{itemize}

Each rubric criterion was written and verified by experts from top PE and IB firms. By tracking the steps taken with respect to these three categories, reviewers can audit both workflow and judgment, irrespective of the final output. For ultra long-horizon tasks that can take an analyst 20 to 30 or more hours, the trajectory is invaluable.

DeepFinance tasks test whether an agent can carry decisions through a long assignment and follow best practices along the way. A source chosen early may affect a forecast, the amount a company can borrow, its valuation, and the final recommendation. When an input changes, every calculation that depends on it must change too. An agent should not be penalized only for upstream decisions that go wrong; the quality of the steps and judgments made along the way should also be evaluated, irrespective of output. This is how DeepFinance captures the trajectory, and graders give partial credit when a model takes the right steps.

The rubric score is the percentage of rubric criteria satisfied, averaged over attempts and then over tasks. Attempts that errored were re-run, so every configuration is scored on 10 successful attempts per task.

\subsection{Quality Assurance}
\label{sec:quality-assurance}

Every task was reviewed by multiple layers of agentic verifiers and by human expert peers with backgrounds at top firms such as Evercore and institutions such as Harvard Business School and Wharton. Expert review and verification established four conditions:

\begin{enumerate}
  \item The task is clear, and the source files contain enough information to complete it.
  \item The reference solution is supported by the evidence.
  \item The rubric checks the requested work and accepts valid solutions.
  \item A failed criterion reflects the agent's work, rather than a faulty reference, grader, or environment issue.
\end{enumerate}

Reviewers traced disputed grades and expert commentary back to the source files and agent traces. This distinguishes model mistakes from evaluation mistakes, so that errors in the evaluation process do not influence conclusions about model capability.

\subsection{Harness}
\label{sec:harness}

Unlike many benchmarks, DeepFinance evaluates each model in the harness it is most likely to be used in (Table~\ref{tab:harness}). Each harness has nearly every native tool enabled, including web search. The one exception is the ask-user tool, which is disabled so that every run can finish on its own. Many evaluations cut off internet access so that agents cannot look up answer keys; DeepFinance does not need to. Every run has its own sandbox, and the answer key and rubric sit in separate, isolated grading environments.

The goal is to test agents in environments representative of those PE and IB professionals use day to day. This yields a more realistic evaluation than a single universal harness that imposes arbitrary or artificial constraints on the models.

\begin{table}[htbp]
  \centering
  \caption{Agent harness used for each model family. All harnesses run on Node.js v22.23.4.}
  \label{tab:harness}
  \begin{tabular}{lll}
    \toprule
    Models & Harness & Version \\
    \midrule
    Claude & claude-code & 2.1.318 \\
    GPT & codex & 0.166.0 \\
    Gemini, Muse, DeepSeek, GLM, Nemotron, Inkling & opencode-ai & 1.19.3 \\
    Kimi & kimi-code & 2.4.1 \\
    Qwen & qwen-code & 0.26.6 \\
    Grok & x.ai & 1.0.70 \\
    Mistral & mistral-vibe & 2.24.3 \\
    \bottomrule
  \end{tabular}
\end{table}

\section{Results}
\label{sec:results}

\begin{figure}[t]
  \centering
  \includegraphics[width=\textwidth]{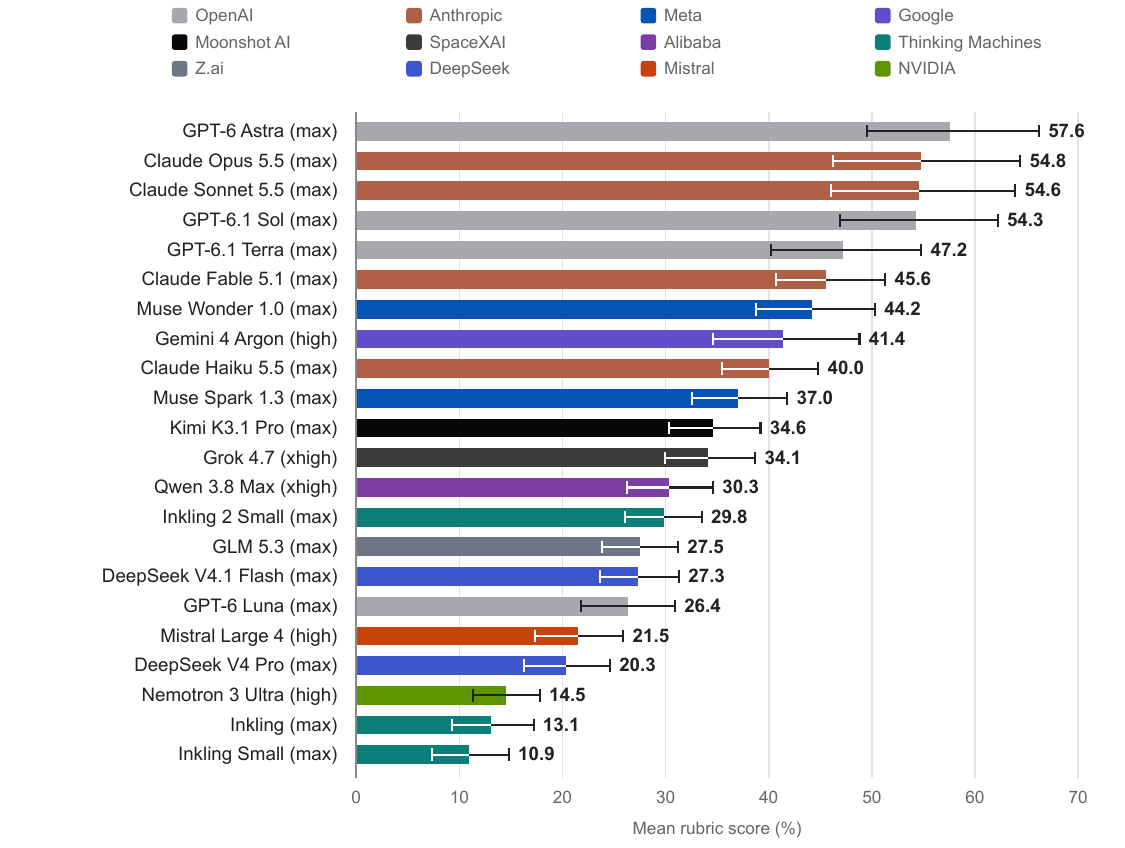}
  \caption{Mean rubric score by model configuration on DeepFinance (21 tasks, 10 attempts per task), colored by lab. Rubric score is the percentage of rubric criteria satisfied. Error bars denote 95\% bootstrap confidence intervals (100,000 resamples over tasks and attempts).}
  \label{fig:leaderboard}
\end{figure}

Figure~\ref{fig:leaderboard} shows the mean rubric score of each configuration, and Table~\ref{tab:results} lists every score with its confidence interval, cost, and token usage.

\begin{table}[htbp]
  \centering
  \caption{Results by model configuration, best first. Cost is mean API cost per completed task; tokens are mean tokens per task.}
  \label{tab:results}
  \small
  \input{results-table}
\end{table}

\subsection{Quantitative Analysis}
\label{sec:quantitative-analysis}

\paragraph{The gap is widening on the frontier.} Anthropic and OpenAI models take the six highest scores. The top four configurations, GPT-6 Astra, Claude Opus 5.5, Claude Sonnet 5.5, and GPT-6.1 Sol (57.6, 54.8, 54.6, and 54.3), are not statistically distinguishable (paired bootstrap, 95\%). The best model from another provider, Muse Wonder 1.0, scores 44.2, and the bottom group (Nemotron 3 Ultra, Inkling, and Inkling Small, at 10.9 to 14.5) sits roughly 45 points below the top.

\begin{figure}[t]
  \centering
  \includegraphics[width=0.85\textwidth]{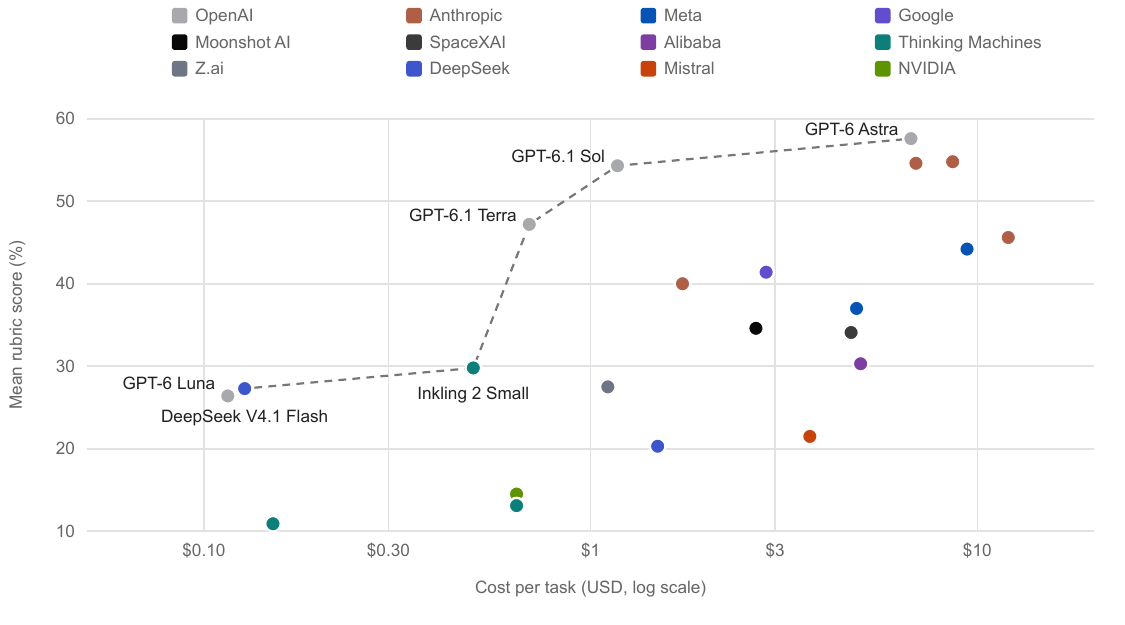}
  \caption{Mean rubric score against mean API cost per task. The dashed line marks the Pareto frontier: configurations not dominated on both score and cost. Frontier configurations are labeled.}
  \label{fig:cost}
\end{figure}

\begin{figure}[t]
  \centering
  \includegraphics[width=0.85\textwidth]{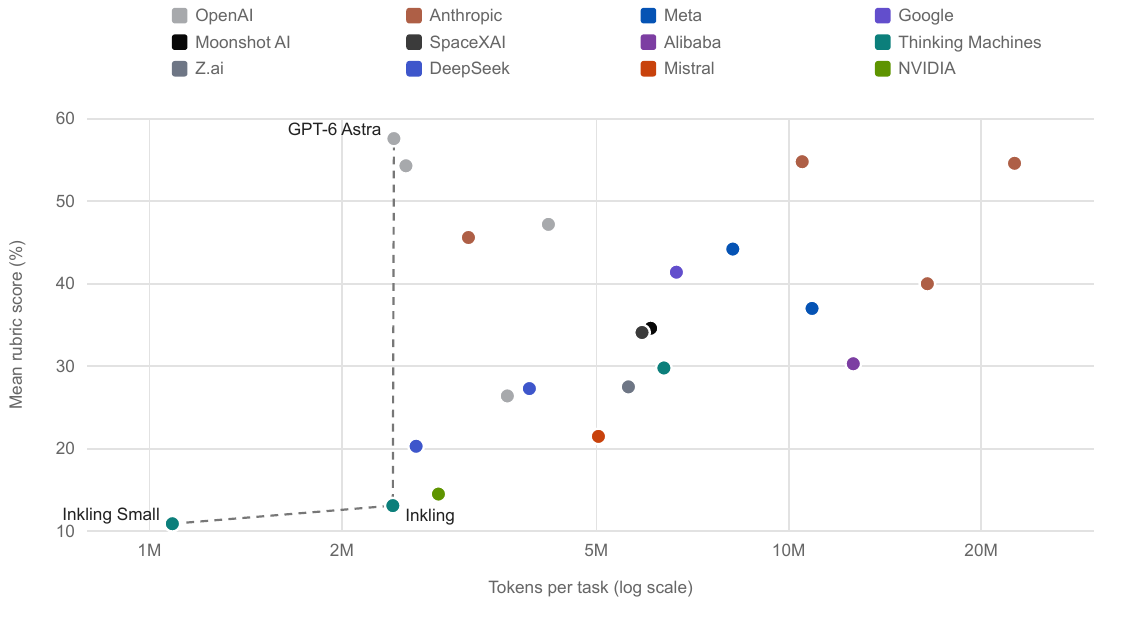}
  \caption{Mean rubric score against mean tokens per task. The dashed line marks the Pareto frontier: configurations not dominated on both score and tokens. Frontier configurations are labeled.}
  \label{fig:tokens}
\end{figure}

\paragraph{GPT-6.1 class models dominate the Pareto frontier.} GPT-6.1 Sol with max reasoning averaged \$1.17 per completed task with a score of 54.3\%. The cheapest model scoring above 30\% was GPT-6.1 Terra, at \$0.69 per completed task with a score of 47.2\%. GPT-6 models make up 4 of the 6 configurations on the cost Pareto frontier across the 22 tested (Figures~\ref{fig:cost} and~\ref{fig:tokens}).

\subsection{Qualitative Analysis}
\label{sec:qualitative-analysis}

\paragraph{Fable for finance: not worth the cost.} Although Claude Fable 5.1 is regarded for its intelligence, this did not carry over to high finance: it scored below GPT-6 Astra, Claude Opus 5.5, Claude Sonnet 5.5, GPT-6.1 Sol, and GPT-6.1 Terra. It was also the most expensive model tested, while scoring under 50\% on end-to-end tasks.

\paragraph{Even frontier models are weak at turning evidence into sound judgment.} In most DeepFinance tasks, agents find the relevant information without recognizing its implications. They also have difficulty challenging a supplied assumption or evaluating better opportunities unless explicitly told to, even where a human analyst would do so instinctively. When a CFO says one thing and an analyst says another, agents make little effort to reconcile the two perspectives.

\paragraph{Upstream mistakes cause cascading downstream failures.} Incorrect earnings, source selection, or completion timing can affect debt capacity, fees, cash flows, valuations, and investor returns. Atomic output checks reveal those consequences, but the number of failed checks should not be interpreted as the number of independent reasoning failures.

\paragraph{Models often execute calculations correctly after selecting the wrong inputs.} In a lending task, agents correctly calculated debt capacity using an incorrect earnings subtotal. In a pharmaceutical royalty task, they correctly applied payment caps and milestone rules to a forecast that omitted an entire sales channel. The agents knew how to perform the calculations but failed to verify the numbers feeding them.

\paragraph{Agents find decision-relevant evidence without acting on it.} In one acquisition task, moving the closing date changed shareholder payouts and lowered the required offer by \$1.25 million. Every attempt by frontier models missed this saving, even though several had identified the buyer's right to choose the date.

\section{Conclusion}
\label{sec:conclusion}

For AI to be useful in dealmaking, its work needs to hold up when someone checks it. Across high-finance assignments that take professionals 20 to 30 hours, agents must choose sound assumptions, use them consistently, and explain the decisions that matter.

DeepFinance evaluates both the result and the work behind it. Auditability gives professionals a way to check the analysis and gives researchers a way to find the mistakes that need fixing. This brings us closer to measuring whether agents can take an assignment from a data room to a decision people can rely on.

\section*{Acknowledgments}

We thank the Julius Research team and the private equity and investment banking professionals who composed and reviewed the DeepFinance tasks.

{\raggedright
}

\clearpage
\appendix

\section{Sample Task Details}
\label{app:sample-task}

This appendix expands the investment due diligence task in Section~\ref{sec:sample-task}. Amounts are in USD millions unless stated.

\subsection{Data Room}
\label{app:data-room}

Table~\ref{tab:data-room} lists the 18 files in the task's data room.

\begin{table}[H]
  \centering
  \caption{Files in the sample task's data room.}
  \label{tab:data-room}
  \begin{tabular}{llr}
    \toprule
    File & Type & Size (KB) \\
    \midrule
    01\_Corliss\_Close\_Ledger.docx & Word document & 37 \\
    02\_Corliss\_Supporting\_Registers.docx & Word document & 46 \\
    03\_Corliss\_Management\_Pack.docx & Word document & 37 \\
    04\_Corliss\_Executed\_Instruments.docx & Word document & 39 \\
    05\_Corliss\_Pricing\_Framework.docx & Word document & 37 \\
    06\_Corliss\_Financial\_Reporting\_Policy.docx & Word document & 36 \\
    07\_Ellery\_Motion\_Borrowings\_20260109.docx & Word document & 37 \\
    08\_Corrib\_Industrial\_Debt\_Schedule\_20260109.docx & Word document & 36 \\
    09\_Corliss\_Share\_Rights\_and\_Option\_Plan.docx & Word document & 37 \\
    10\_Corliss\_Securities\_Register.docx & Word document & 37 \\
    11\_Corliss\_Sale\_Mandate\_and\_Costs.docx & Word document & 37 \\
    12\_Corliss\_IC\_Price\_Check\_Template.xlsx & Spreadsheet & 6 \\
    13\_Corliss\_Working\_20260106.xlsx & Spreadsheet & 6 \\
    14\_Email\_Re\_warranty\_costs.eml & Email & 1 \\
    15\_Email\_Whitlow\_invoice.eml & Email & 1 \\
    16\_Email\_Q4\_register\_D-284\_dates.eml & Email & 1 \\
    17\_Investment\_Team\_Chat\_Export.txt & Text file & 1 \\
    18\_Email\_Quarrie.eml & Email & 1 \\
    \bottomrule
  \end{tabular}
\end{table}

\subsection{Reference Solution}
\label{app:reference-solution}

The reference solution is a workbook, \texttt{Corliss\_IC\_Price\_Check.xlsx}, with nine sheets: Price check, Allocation, Workings, Completion date, Structures, Assumptions, Waterfall, Inputs, and Audit checks. Table~\ref{tab:reference-solution} summarizes the Price check sheet.

\begin{table}[H]
  \centering
  \caption{Key figures from the reference solution's Price check sheet (USD millions).}
  \label{tab:reference-solution}
  \begin{tabular}{lr}
    \toprule
    Item & Amount \\
    \midrule
    Reconciled FY2025 revenue & 42.880 \\
    Reconciled FY2025 EBITDA & 7.180 \\
    Sustainable EBITDA & 7.520 \\
    Peer median EV/EBITDA & 8.735x \\
    Enterprise value & 65.684 \\
    Net debt & 11.290 \\
    Working capital adjustment & (0.517) \\
    Equity value & 53.877 \\
    \midrule
    Price to put to IC (all cash, completion April 1, 2026) & 50.803 \\
    Hale group cash at completion at that price & 18.500 \\
    Declan's structure: price needed at completion (April 1) & 50.603 \\
    Declan's structure: expected cost, with the 3.00 Quarrie payment at 76.95\% & 52.911 \\
    \bottomrule
  \end{tabular}
\end{table}

Two judgments drive the answer. First, sustainable EBITDA includes nothing for Quarrie: both parts need the same Supplier Quality Board approval (95\%) and then their own separate field trials (90\% each), so the chance both enter production is $95\% \times 90\% \times 90\% = 76.95\%$, below the pricing framework's 80\% threshold. Second, the sale mandate lets the buyer nominate the completion date. Nominating April 1, 2026, after the third anniversary of the Series A issue, lowers the price needed to meet the Hale condition (18.500 in cash at completion for the Hale group) from 52.058 to 50.803. The reference solution's recommendation reads: ``Offer 50.803 in cash with no Quarrie payment, nominating completion on April 1, 2026 (any business day in April up to the long-stop). Declan's structure costs more on the framework's probability-weighted basis on either date.''

\subsection{Rubric}
\label{app:rubric}

Table~\ref{tab:rubric} shows 9 of the task's 136 rubric criteria, three from each category.

\begin{table}[H]
  \centering
  \caption{Sample rubric criteria for the sample task (9 of 136 shown).}
  \label{tab:rubric}
  \begin{tabular}{lp{0.78\textwidth}}
    \toprule
    Category & Criterion \\
    \midrule
    Output & The workbook reports reconciled FY2025 revenue of 42.880. \\
    Output & The workbook reports sustainable EBITDA of 7.51 to 7.53. \\
    Output & The workbook puts a price of 50.80 to 50.81 to IC. Justifiable alternative: 50.803 or 50.804; the exact price is 50.8034 and 50.804 is the three-decimal price that clears 18.500. \\
    \midrule
    Workflow & The workbook shows a line-by-line bridge from booked revenue to reconciled revenue, with the record behind each line. \\
    Workflow & The workbook calendarizes Ellery's earnings: the year to March 2025, less the nine months to December 2024, plus the nine months to December 2025. \\
    Workflow & The workbook counts the shared EC-22 approval once, times 90\% for each part's own field trial. \\
    \midrule
    Judgment & The workbook treats the equity value, not Owen's 56, as the most Pinecrest may pay. \\
    Judgment & The workbook recommends all cash with no Quarrie payment. \\
    Judgment & The warranty row leaves the IC price whole, with no deduction. \\
    \bottomrule
  \end{tabular}
\end{table}

\end{document}

%% file: results-table.tex
\begin{tabular}{rlllrrr}
\toprule
\# & Model (reasoning) & Lab & Score (\%) & 95\% CI & Cost (USD) & Tokens \\
\midrule
1 & GPT-6 Astra \reasoning{max} & OpenAI & 57.6 & 49.5--66.2 & 6.73 & 2.41M \\
2 & Claude Opus 5.5 \reasoning{max} & Anthropic & 54.8 & 46.2--64.4 & 8.63 & 10.49M \\
3 & Claude Sonnet 5.5 \reasoning{max} & Anthropic & 54.6 & 46.0--63.9 & 6.94 & 22.55M \\
4 & GPT-6.1 Sol \reasoning{max} & OpenAI & 54.3 & 46.9--62.2 & 1.17 & 2.52M \\
5 & GPT-6.1 Terra \reasoning{max} & OpenAI & 47.2 & 40.2--54.8 & 0.69 & 4.21M \\
6 & Claude Fable 5.1 \reasoning{max} & Anthropic & 45.6 & 40.7--51.3 & 12.02 & 3.15M \\
7 & Muse Wonder 1.0 \reasoning{max} & Meta & 44.2 & 38.8--50.3 & 9.41 & 8.18M \\
8 & Gemini 4 Argon \reasoning{high} & Google & 41.4 & 34.6--48.8 & 2.84 & 6.67M \\
9 & Claude Haiku 5.5 \reasoning{max} & Anthropic & 40.0 & 35.5--44.8 & 1.73 & 16.47M \\
10 & Muse Spark 1.3 \reasoning{max} & Meta & 37.0 & 32.6--41.8 & 4.87 & 10.87M \\
11 & Kimi K3.1 Pro \reasoning{max} & Moonshot AI & 34.6 & 30.3--39.2 & 2.67 & 6.08M \\
12 & Grok 4.7 \reasoning{xhigh} & SpaceXAI & 34.1 & 29.9--38.7 & 4.71 & 5.90M \\
13 & Qwen 3.8 Max \reasoning{xhigh} & Alibaba & 30.3 & 26.3--34.6 & 4.99 & 12.62M \\
14 & Inkling 2 Small \reasoning{max} & Thinking Machines & 29.8 & 26.1--33.5 & 0.50 & 6.38M \\
15 & GLM 5.3 \reasoning{max} & Z.ai & 27.5 & 23.8--31.2 & 1.11 & 5.61M \\
16 & DeepSeek V4.1 Flash \reasoning{max} & DeepSeek & 27.3 & 23.6--31.3 & 0.13 & 3.93M \\
17 & GPT-6 Luna \reasoning{max} & OpenAI & 26.4 & 21.8--30.9 & 0.12 & 3.63M \\
18 & Mistral Large 4 \reasoning{high} & Mistral & 21.5 & 17.3--25.9 & 3.69 & 5.04M \\
19 & DeepSeek V4 Pro \reasoning{max} & DeepSeek & 20.3 & 16.3--24.6 & 1.49 & 2.61M \\
20 & Nemotron 3 Ultra \reasoning{high} & NVIDIA & 14.5 & 11.3--17.8 & 0.64 & 2.83M \\
21 & Inkling \reasoning{max} & Thinking Machines & 13.1 & 9.3--17.2 & 0.64 & 2.40M \\
22 & Inkling Small \reasoning{max} & Thinking Machines & 10.9 & 7.3--14.8 & 0.15 & 1.09M \\
\bottomrule
\end{tabular}